%% file: main.tex
\documentclass{article}

\usepackage{graphicx} 
\usepackage[a4paper, top=2.5cm, bottom=2.5cm, left=2cm, right=2cm]{geometry}
\usepackage{booktabs}
\usepackage{multirow}
\usepackage{amsmath} 
\usepackage{amssymb} 
\usepackage{verbatim}
\usepackage{subcaption}
\usepackage{float}
\usepackage{tabularx}
\usepackage{xcolor}
\usepackage{tikz}
\usetikzlibrary{arrows.meta, positioning}
\usepackage{colortbl}  

\usepackage{hyperref}
\hypersetup{
    colorlinks=true,
    linkcolor=blue,
    filecolor=magenta,      
    urlcolor=blue,
}

\definecolor{darkgreen}{rgb}{0.0, 0.5, 0.0}

\title{SF20K Competition 2025: Summary and findings}
\author{
    Ridouane Ghermi,
    Xi Wang,
    Vicky Kalogeiton,
    Ivan Laptev
}
\date{}

\begin{document}

\maketitle

\begin{center}
    Project: \url{https://ridouaneg.github.io/sf20k.html} \\
    Competition: \url{https://huggingface.co/spaces/SLoMO-Workshop/SF20KCompetition} \\
    Dataset: \url{https://huggingface.co/datasets/rghermi/sf20k}
\end{center}

\begin{figure}[H] 
    \centering
    \includegraphics[width=\linewidth]{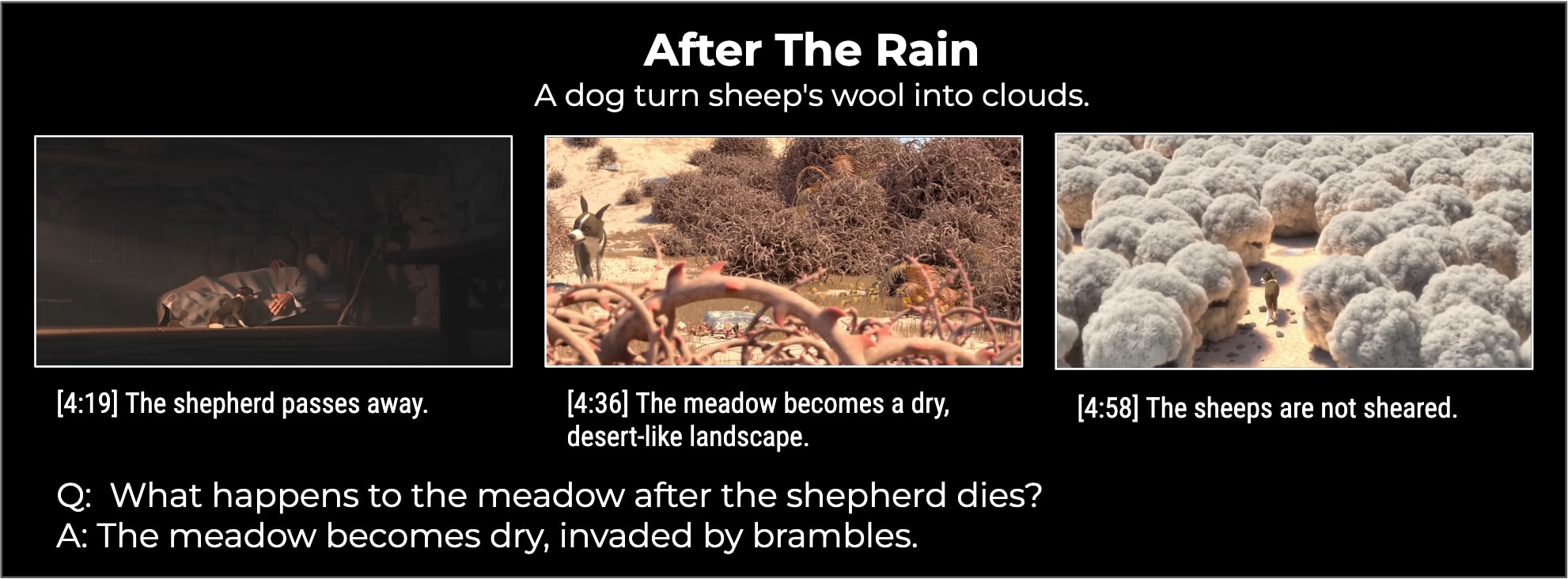}
    \caption{\textbf{A sample from the SF20K competition}, including the movie title, a sentence describing the story, a few frames with their corresponding captions, a question and a ground-truth answer. In this example, we display the specific timestamps that enables to answer the question.}
    \label{fig:qa_example}
\end{figure}

\begin{abstract}
This report presents the results and findings of the first edition of the Short-Films 20K (SF20K) Competition, held in conjunction with the SLoMO Workshop\footnote{\url{https://slomo-workshop.github.io/}} at ICCV 2025. The competition is designed to advance story-level video understanding beyond short-clip action recognition, introducing an open-ended video question-answering task built on a corpus of amateur short films. This setup ensures that models must rely on multimodal understanding rather than memorization of popular movies. Evaluation is conducted using the SF20K-Test benchmark (95 movies, 979 question-answer pairs) and scored via LLM-QA-Eval, an automated judge based on GPT-4.1-nano. The competition attracted 22 teams and 286 submissions across two tracks: a Main Track with unrestricted model size and a Special Track limited to models under 8 billion parameters. The winning team achieved 65.7\% accuracy on the Main Track and 48.7\% on the Special Track, against a human performance ceiling of 91.7\%. Our analysis reveals several key findings: narrative-aware, shot-level processing consistently outperforms uniform frame sampling; well-designed multi-stage pipelines using smaller models can match or exceed end-to-end inference with models over 30$\times$ larger; and subtitle quality is a dominant factor in performance. These results highlight that the primary bottleneck in long-form video QA lies in information selection and reasoning structure rather than raw model capacity, and that a substantial gap remains between current methods and human-level narrative comprehension.
%
%
\end{abstract}


\section{Introduction}

The field of video-language understanding has seen rapid progress in short-clip action recognition, yet existing benchmarks remain largely limited to brief videos with simple narratives. Movie-based alternatives, while offering longer and more complex content, often suffer from saturation or ‘‘memorization,'' where large language models answer questions about popular films using parametric knowledge rather than genuine visual comprehension. These limitations highlight a critical gap: the lack of benchmarks that require models to move beyond frame-level feature extraction toward holistic, narrative-driven reasoning—encompassing visual storytelling, character identification across scenes, and long-range causal understanding.

The Short-Films 20K (SF20K) Competition was established to address this gap. Built on a novel corpus of 20,143 amateur short films~\cite{sf20k}, the competition introduces a challenging open-ended question-answering task that ensures models must rely on active multimodal grounding rather than pretrained internal knowledge. Evaluation focuses on the SF20K-Test benchmark~\cite{sf20k}, which comprises 95 movies and 979 manually crafted question-answer pairs, with an average film duration of 12 minutes. This design specifically targets the understanding of complex, story-level narratives that unfold over extended timescales. We provide a sample from the benchmark in Figure~\ref{fig:qa_example}.



\section{Challenge setup}

\subsection{Task}

The competition focuses on Open-Ended Video Question Answering (VideoQA), requiring models to generate concise, free-form natural language responses to questions about short films. This task demands narrative-level comprehension: models must identify characters across scenes, track causal chains, and synthesize information spanning the full duration of each film.

\subsection{Submission policy}

To explore the trade-off between raw performance and efficient design, the competition is divided into two tracks. The Main Track places no constraints on model size or compute, while the Special Track limits models to under 8 billion parameters, encouraging the development of efficient and innovative strategies.

The competition is hosted via a Hugging Face Space~\footnote{\url{https://huggingface.co/spaces/SLoMO-Workshop/SF20KCompetition}} and proceeds in two phases. During Phase 1 (Public), participants can submit up to three predictions per day to a public test set to iteratively refine their methods. Phase 2 (Private) determines the final rankings from a single submission on a held-out private test set. To ensure the evaluation remains immune to data contamination, the private set consists of movies released during the competition period itself (July to October 2025).


\input{figures/metric_def}

\subsection{Evaluation metric}

To mitigate the limitations of traditional n-gram metrics such as CIDEr~\cite{cider} and BLEU~\cite{bleu}, which struggle to capture semantic similarity in open-ended responses, we adopt LLM-QA-Eval~\cite{sf20k}. This automated metric uses GPT-4.1-Nano~\cite{gpt41} as a judge via the official OpenAI API. Given a question, a predicted response, and the ground-truth answer, the judge assigns a binary correctness label and a qualitative score from 0 to 5. The final leaderboard is ranked by average correctness, providing a robust measure of a model's ability to answer the narrative-driven questions. We illustrate the metric definition in Figure~\ref{fig:metric_def}.


\section{Baseline performance}


In Table~\ref{tab:baselines}, we establish a performance overview using a series of baseline configurations. A ‘‘Blind guessing''
baseline (question-only, i.e., GPT-5-mini without any context information) achieves only 14.7\% accuracy, confirming that the answers cannot be guessed without watching the movie. Our primary baseline, Qwen2.5-VL-7B~\cite{qwen25vl}, achieves 30.7\% accuracy when sampling 512 frames, while the state-of-the-art method, Gemini-2.5-Pro~\cite{gemini25}, reached 59.7\%. The significant gap between the SOTA and Human Performance (91.7\%) demonstrates that there is room for improvement and highlights the difficulty of narrative understanding from current methods.

\input{tables/baselines}
\input{figures/ablations}

\noindent\textbf{Ablation study.} Our experiments demonstrate that the fusion of visual data and subtitles is vital for success on the benchmark. A visual-only configuration achieves only 20.8\% accuracy, from 30.7\% for the full vision-language setup, whereas a language-only model (using only subtitles as inputs) reaches 27.8\% (see Figure~\ref{fig:perf_vs_modality}). Moreover, performance increases with the number of input frames (see Figure~\ref{fig:perf_vs_nframes}), peaking at 512 frames. These results suggest that the best methods will likely need to handle rich context, with exhaustive input frames and multimodal inputs.

\section{Competition results}

\subsection{Participants}

The SF20K Competition ran from July 1st to October 10th, 2025, attracting significant interest from the multimodal research community, with a total of 22 active teams participating. Table~\ref{tab:solutions} summarizes the solutions submitted by all teams.


\subsection{Submissions}

Over the course of the competition, teams contributed 286 valid submissions across both tracks. Figure~\ref{fig:perf_evo} tracks the progress of each team throughout the competition period. Table~\ref{tab:leaderboard} presents the final leaderboard for both the Main Track (unrestricted model size) and the Special Track ($<$8B parameters). Rankings were determined by accuracy on the private test set, which was constructed from films released \emph{during} the competition period (July to October 2025) to prevent any form of data contamination.

Team WXYZ won both tracks, achieving 65.7\% on the Main Track and 48.7\% on the Special Track. The rankings shifted substantially between the public and private phases. For instance, team R\&T ranked 1st on the public Main Track leaderboard but did not place in the top 3 on the private set, validating the competition's anti-overfitting design based on temporally disjoint test films. The gap between the best-performing system (65.7\%) and human performance (91.7\%) remains substantial, underscoring the difficulty of long-form narrative video understanding.



\input{figures/perf_through_time}
\input{tables/final_rankings}

\subsection{Key findings}

%

Below, we highlight the main findings that emerge from a cross-team analysis of the submitted approaches.

\noindent\textbf{Narrative-aware sampling outperforms uniform sampling.}
The most striking finding is the advantage of \emph{shot-level, narrative-aware} processing over standard uniform frame sampling. The winning WXYZ team structured their entire pipeline around video shots as fundamental reasoning units. Their system consists of four modules: a two-stage keyframe selection process,
 a narrative rhythm detection module that estimates story pacing from audio energy, speaker speed, shot duration, and optical flow, shot-level story description agents that iteratively generate and refine narrative descriptions, and a multi-stage chain-of-thought reasoning module with LLM-based answer verification. This approach outperformed much larger models used in simpler configurations: WXYZ achieved 65.7\% using Qwen2.5-VL-7B~\cite{qwen25vl} (7B parameters) as their core VLM in the Main Track, comparable to sudook's 65.3\% using Qwen3-VL-235B~\cite{qwen3}, a model over 30$\times$ larger. This suggests that \emph{how} visual and textual information is organized and presented to the model matters as much as, or more than, the model's raw capacity. 
 
\noindent\textbf{Multi-stage pipelines can compensate for model size.}
More broadly, well-designed multi-stage pipelines using smaller models matched or exceeded end-to-end inference with much larger models. Team BASELINE adopted a ``perceive then reason'' paradigm, separating visual perception (Qwen2.5-VL-7B~\cite{qwen25vl} for frame-level captioning) from reasoning (GPT-5~\cite{gpt5} for final answer generation), achieving 3rd place in the Main Track. Team eloral-wxy similarly achieved 2nd place in the Special Track with a fully zero-shot, API-based approach using GLM~\cite{glm} for video summarization and GPT~\cite{chatgpt} for confidence-based answer refinement. These findings suggest that the bottleneck in long-form video QA lies less in model capacity and more in information selection and reasoning structure.
 
\noindent\textbf{Subtitle quality is a dominant factor.}
Across all teams, high-quality subtitle extraction emerged as one of the most impactful components. The competition provided official subtitle files (extracted using Whisper-tiny~\cite{whisper}), but several teams invested in generating improved subtitles using ASR models. Teams WXYZ and eg both employed dual ASR systems (Whisper~\cite{whisper} and Parakeet~\cite{parakeet}), selecting or fusing outputs depending on audio conditions. Team R\&T demonstrated that replacing official subtitles with custom Whisper-large-v3~\cite{whisper} extractions (with deduplication) boosted Keye-VL-1.5-8B~\cite{keyevl} from 44.2\% to 46.4\% and Qwen3-VL-30B~\cite{qwen3vl} from 49.6\% to 55.0\% on the public benchmark. These findings confirm that audio-derived linguistic context is critical for narrative understanding.

\input{tables/solutions}
 

\noindent\textbf{Dense frame sampling requires better temporal encoding.}
%
%
Simply increasing the number of input frames does not reliably improve performance. Team R\&T observed that for Qwen2.5-VL-7B~\cite{qwen25vl}, accuracy slightly dropped from 41.6\% at 128 frames to 38.1\% at 256 frames, suggesting that standard models struggle to process redundant visual information effectively. Notably, models with specialized temporal encoding, such as Keye-VL's slow-fast architecture or Qwen3-VL's MRoPE-Interleave, exhibited less performance degradation at higher frame counts, suggesting that architectural innovations in temporal models are needed to scale visual context effectively.

\noindent\textbf{Fine-tuning on limited data provides modest gains.}
Several teams experimented with fine-tuning on the SF20K~\cite{sf20k} training set, though in all cases with limited data (500 samples). Team eg applied LoRA to Qwen2.5-VL-7B~\cite{qwen25vl} using category-balanced multiple-choice questions and demonstrated that MCQA fine-tuning transfers to open-ended QA. Teams WXYZ and sudook (Special Track) also performed fine-tuning on 500 training samples, reporting moderate improvements. However, the core architectural innovations (shot-aware pipelines, narrative rhythm detection, chain-of-thought verification) contributed more substantially to final performance than fine-tuning alone. That said, this may be explained by the limited computational resources available to participants, and large-scale task-specific training could yield different conclusions.

\noindent\textbf{The human-machine gap remains large.}
Despite the progress made during the competition, the best system accuracy (65.7\%) remains 26 points below human performance (91.7\%). This gap is especially pronounced in the Special Track, where the best model reached only 48.7\%. The remaining challenges involve long-range causal reasoning, character identification across distant scenes, and the integration of subtle visual cues (facial expressions, environmental changes) with dialog, all of which require deeper multimodal understanding than current architectures provide.

\section{Conclusion}
 
The first edition of the SF20K Competition successfully brought together the multimodal research community around the challenge of story-level video understanding. The diversity of submitted approaches, from large-scale end-to-end models to lightweight training-free pipelines, demonstrated that this task requires innovations across multiple axes: temporal reasoning, information selection, subtitle quality, and narrative structure modeling. The consistent finding that pipeline design and data quality matter as much as model scale offers an encouraging direction for the field, particularly for resource-constrained settings. The substantial gap to human performance ensures that SF20K will continue to serve as a challenging and informative benchmark for future research in long-form video comprehension.


\noindent\textbf{Acknowledgments.} This work was supported by ANR-22-CE23-0007, a Hi!Paris grant and fellowship, and the ANR/France 2030 program (ANR-23-IACL-0005). Additional support was provided by the Institute of Information \& Communications Technology Planning \& Evaluation (IITP), funded by the Korean Government (MSIT) under Grant No.\ RS-2024-00457882 (National AI Research Lab Project). We would like to thank Junyu Xie, Tengda Han, Max Bain, Arsha Nagrani, Gül Varol, Weidi Xie, and Andrew Zisserman for the valuable discussions around the competition, and for helping with the annotation of the private test set.



\bibliographystyle{plain}
\bibliography{main}

\end{document}

%% file: figures/metric_def.tex
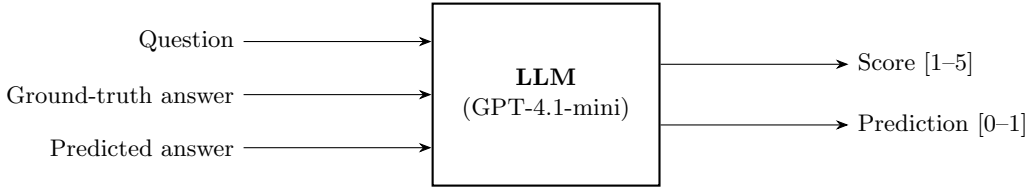
\begin{figure}[t]
\centering
\begin{tikzpicture}[
    >=Stealth,
    every node/.style={font=\small},
    box/.style={draw, thick, minimum width=3cm, minimum height=2.4cm, align=center}
]

\node (llm) [box] at (4, 0) {\textbf{LLM}\\(GPT-4.1-mini)};

\node[anchor=east] (q)  at (0,  0.7) {Question};
\node[anchor=east] (gt) at (0,  0.0) {Ground-truth answer};
\node[anchor=east] (pa) at (0, -0.7) {Predicted answer};

\node[anchor=west] (score) at (8,  0.4) {Score [1--5]};
\node[anchor=west] (pred)  at (8, -0.4) {Prediction [0--1]};

\draw[->] (q.east)  -- ([yshift=0.7cm]  llm.west);
\draw[->] (gt.east) -- (llm.west);
\draw[->] (pa.east) -- ([yshift=-0.7cm] llm.west);

\draw[->] ([yshift=0.4cm]  llm.east) -- (score.west);
\draw[->] ([yshift=-0.4cm] llm.east) -- (pred.west);

\end{tikzpicture}
\caption{\textbf{The LLM-QA-Eval metric.} Given a question, the metric uses an LLM (i.e., GPT-4.1-nano) to compare a predicted answer to the ground-truth answer, assigning a 1-5 score and a 0-1 correctness label. The final metric is the average correctness across all $N=979$ samples.}
\label{fig:metric_def}
\end{figure}

%% file: tables/baselines.tex
\begin{table}[tb]
  \centering
  \begin{tabular}{@{}lllll@{}}
    \toprule
    Method name & Model & \#Frames & Subtitles & Private Acc. (\%) \\
    \midrule
    Random & - & - & - & 0.00 \\
    Blind guessing & GPT-5-mini~\cite{gpt5} & 0 & $\times$ & 14.7 \\
    Baseline & Qwen2.5-VL-7B~\cite{qwen25vl} & 512 & $\checkmark$ & 30.7 \\
    State-of-the-art & Gemini-2.5-Pro~\cite{gemini25} & 512 & $\checkmark$ & 59.7 \\
    Human performance & - & - & - & 91.7 \\
    Oracle & - & - & - & 100.0 \\
  \bottomrule
  \end{tabular}
  \caption{\textbf{Baseline performance.} \textit{Random} corresponds to the metric's lowest possible score, while \textit{Oracle} to the theoretical maximum. \textit{Human performance} is a realistic upper-bound derived from manual annotations. \textit{Blind guessing} uses an LLM to guess the answer based on the question only, as a sanity check, while \textit{Baseline} and \textit{State-of-the-art} corresponds to recent end-to-end vision-language models. The subtitles are extracted using \textit{Whisper-tiny}.
  }
  \label{tab:baselines}
\end{table}

%% file: figures/ablations.tex
\begin{figure}[t]
  \centering
  \begin{subfigure}[b]{0.48\textwidth}
    \centering
    \includegraphics[width=\textwidth]{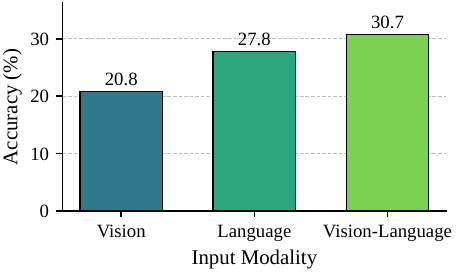}
    \caption{Input modality}
    \label{fig:perf_vs_modality}
  \end{subfigure}
  \hfill
  \begin{subfigure}[b]{0.48\textwidth}
    \centering
    \includegraphics[width=\textwidth]{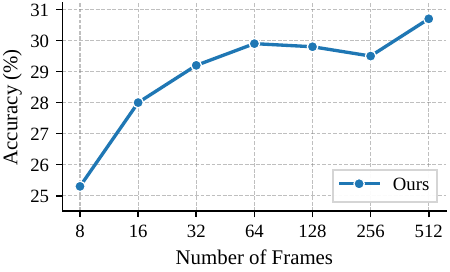}
    \caption{Number of input frames}
    \label{fig:perf_vs_nframes}
  \end{subfigure}
  \caption{\textbf{Ablation study.} Qwen2.5-VL-7B's performance depending on the input modality (left) and number of input frames (right).}
  \label{fig:ablations}
\end{figure}

%% file: figures/perf_through_time.tex
\begin{figure}[t]
\centering
\includegraphics[width=\textwidth]{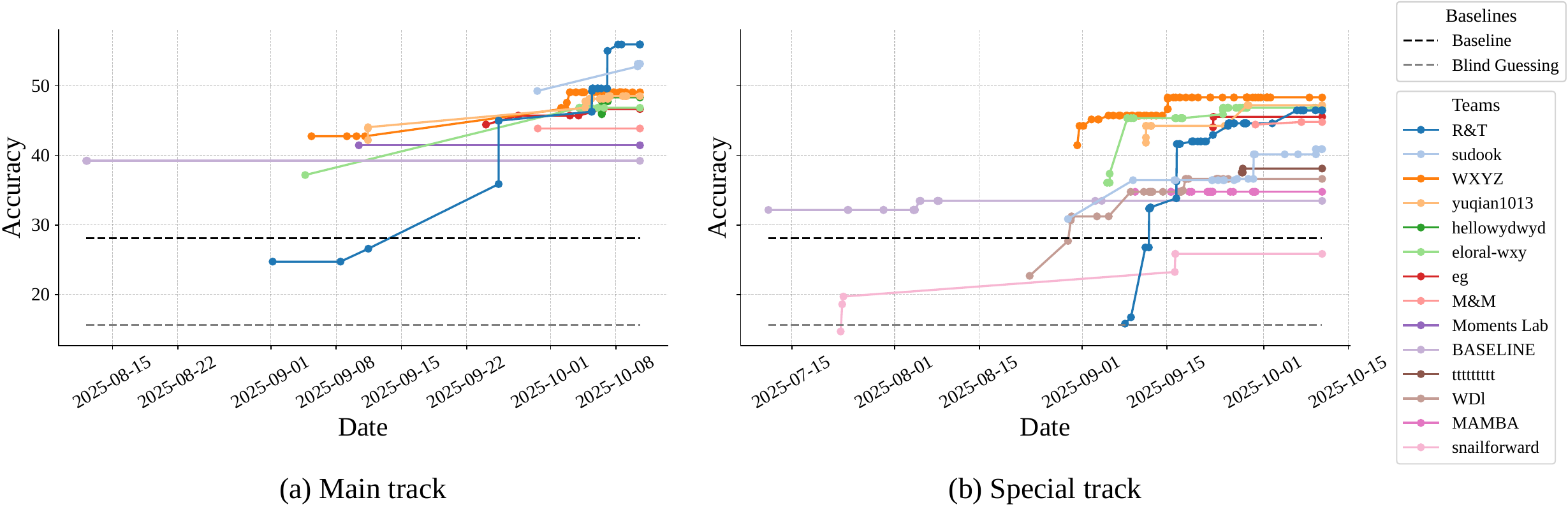}
\caption{\textbf{Teams' performance on the public leaderboard over the course of the competition.}}
\label{fig:perf_evo}
\end{figure}

%% file: tables/final_rankings.tex
\begin{table}[tb]
    \centering
    \begin{tabular}{lrccc}
        \toprule
         Track & Rank &  Team Name &  Public Acc. (\%) &  Private Acc. (\%) \\
        \midrule
        \multirow{5}{*}{Main}
        & 1 & WXYZ & 49.0 & 65.7 \\
        & 2 & sudook & 53.1 & 65.3 \\
        & 3 & BASELINE & 40.8 & 59.8\\
        & 4 & R\&T & 55.9 & 58.5 \\
        & 5 & eg & 46.6 & 53.5 \\
        \midrule
        \multirow{5}{*}{Special}
        & 1 & WXYZ & 48.3 & 48.7 \\
        & 2 & eloral-wxy & 47.2 & 46.2 \\
        & 3 & sudook & 39.2 & 44.8 \\
        & 4 & R\&T & 46.4 & 42.4 \\
        & 5 & eg & 45.5 & - \\
        \bottomrule
    \end{tabular}
    \caption{\textbf{Final leaderboard for the SF20K Competition.} Teams are ranked based on their performance on the Private Test set. The gap between Public and Private scores highlights the models' generalization to unreleased content.}
    \label{tab:leaderboard}
\end{table}

%% file: tables/solutions.tex
\begin{table*}[t]
    \centering
    \small
    \begin{tabular}{p{1.6cm}p{3.5cm}p{8.2cm}cp{1.0cm}}
    \toprule
    \textbf{Team} & \textbf{Model(s)} & \textbf{Key Technique} & \textbf{FT} & \textbf{Private Acc.} \\
    \midrule
    \rowcolor{gray!20} \multicolumn{5}{l}{Main Track} \\
    \addlinespace
    WXYZ & Qwen2.5-VL-7B~\cite{qwen25vl}, Qwen2.5-7B~\cite{qwen25}, Gemini-2.5-Flash~\cite{gemini25} & Shot-aware keyframe selection, narrative rhythm detection, story description agents, multi-stage CoT with LLM verification & \checkmark & 65.7 \\
    \addlinespace
    sudook & Qwen3-VL-235B-A22B-Thinking~\cite{qwen3vl} & End-to-end inference via DashScope API with prompt-engineered frame + subtitle integration & \texttimes & 65.3 \\
    \addlinespace
    BASELINE & Qwen2.5-VL-7B~\cite{qwen25vl}, GPT-5~\cite{gpt5} & Three-stage: Whisper~\cite{whisper} ASR $\rightarrow$ VLM visual tagging $\rightarrow$ LLM reasoning & \texttimes & 59.8 \\
    \addlinespace
    R\&T & Keye-VL-1.5-8B~\cite{keyevl}, Qwen3-VL-30B~\cite{qwen3vl} & Extensive model benchmarking; custom Whisper subtitle extraction with deduplication & \texttimes & 58.5 \\
    \addlinespace
    \rowcolor{gray!20} \multicolumn{5}{l}{Special Track} \\
    \addlinespace
    WXYZ & Qwen2.5-VL-7B~\cite{qwen25vl}, Qwen2.5-7B~\cite{qwen25} & Shot-aware pipeline with LLM-VLM iterative thinking chain & \checkmark & 48.7 \\
    \addlinespace
    sudook & InternVL3.5-8B~\cite{internvl35} & Dynamic high-resolution, Cascade RL, Visual Resolution Router & \checkmark & 44.9 \\
    \addlinespace
    eloral-wxy & GLM~\cite{glm} (API) & Prompt-engineered pipeline: summarize $\rightarrow$ QA $\rightarrow$ GPT confidence-based refinement & \texttimes & 46.2 \\
    \addlinespace
    eg & Qwen2.5-VL-7B~\cite{qwen25vl} & LoRA fine-tuning on 500 category-balanced MCQA; dual ASR (Whisper~\cite{whisper} + Parakeet~\cite{parakeet}) & \checkmark & -- \\
    \bottomrule
    \end{tabular}
    \caption{\textbf{Summary of submitted solutions.} ``FT'' indicates whether fine-tuning was performed.}
    \label{tab:solutions}
\end{table*}